\documentclass[
]{ceurart}

\usepackage{listings}
\usepackage{graphicx}
\usepackage{amsmath,amssymb,amsfonts}
\usepackage{algorithmic}
\usepackage[ruled]{algorithm2e}
\usepackage{textcomp}
\usepackage{tabularx}
\usepackage{booktabs}
\usepackage{subcaption} 
\usepackage{caption}
\usepackage{cleveref}
\usepackage{wrapfig} 
\begin{document}

\copyrightyear{2024}
\copyrightclause{Copyright for this paper by its authors.
  Use permitted under Creative Commons License Attribution 4.0
  International (CC BY 4.0).}

\conference{TempXAI@ECML-PKDD'24: Explainable AI for Time Series and Data Streams Tutorial-Workshop, Sep. 9\textsuperscript{th}, 2024, Vilnius, Lithunia}
\title{Interpreting Outliers in Time Series Data through Decoding Autoencoder}

\author[1,2]{Patrick Knab}
\author[1]{Sascha Marton}
\author[1]{Christian Bartelt}
\author[2]{Robert Fuder}

\address[1]{University of Mannheim, Germany}
\address[2]{Robert Bosch GmbH, Bühl, Germany}
\begin{abstract}
  Outlier detection is a crucial analytical tool in various fields. In critical systems like manufacturing, malfunctioning outlier detection can be costly and safety-critical. Therefore, there is a significant need for explainable artificial intelligence (XAI) when deploying opaque models in such environments.
    This study focuses on manufacturing time series data from a German automotive supply industry. We utilize autoencoders to compress the entire time series and then apply anomaly detection techniques to its latent features. 
    For outlier interpretation, we \textit{\textbf{i)}} adopt widely used XAI techniques to the autoencoder's encoder. Additionally, \textit{\textbf{ii)}} we propose \textbf{AEE}, \textbf{A}ggregated \textbf{E}xplanatory \textbf{E}nsemble, a novel approach that fuses explanations of multiple XAI techniques into a single, more expressive interpretation. For evaluation of explanations, \textit{\textbf{iii)}} we propose a technique to measure the quality of encoder explanations quantitatively. Furthermore, we qualitatively assess the effectiveness of outlier explanations with domain expertise.
\end{abstract}


\maketitle

\section{Introduction} \label{Intro}


Outliers represent exceptional instances that differ from a normal data distribution \cite{10.1145/3381028}. Artificial intelligence (AI) is pivotal in outlier (anomaly) detection applications, particularly in domains with high-dimensional data, such as time series. 
By analyzing patterns, trends, and dependencies, algorithms can effectively identify outliers and anomalous events in various domains, ranging from finance and healthcare to industrial processes \cite{2, 10.1145/3381028, SHAP_AE}. In particular, manufacturing processes generate vast amounts of time series data, making timely and accurate outlier detection critical for maintaining operational efficiency and safety. However, opaque neural networks (NN) often lack the interpretability necessary for high-stakes environments \cite{SHAP_AUTO, SHAP_AE}. Consequently, explaining the model's decisions through explainable artificial intelligence (XAI) is essential to provide transparency and foster trust in automated decision-making \cite{molnar2019, ho2024multivariatetimeseriesanomalydetection, s23052844}.

\begin{figure}[ht]
\centering
\includegraphics[width=\linewidth]{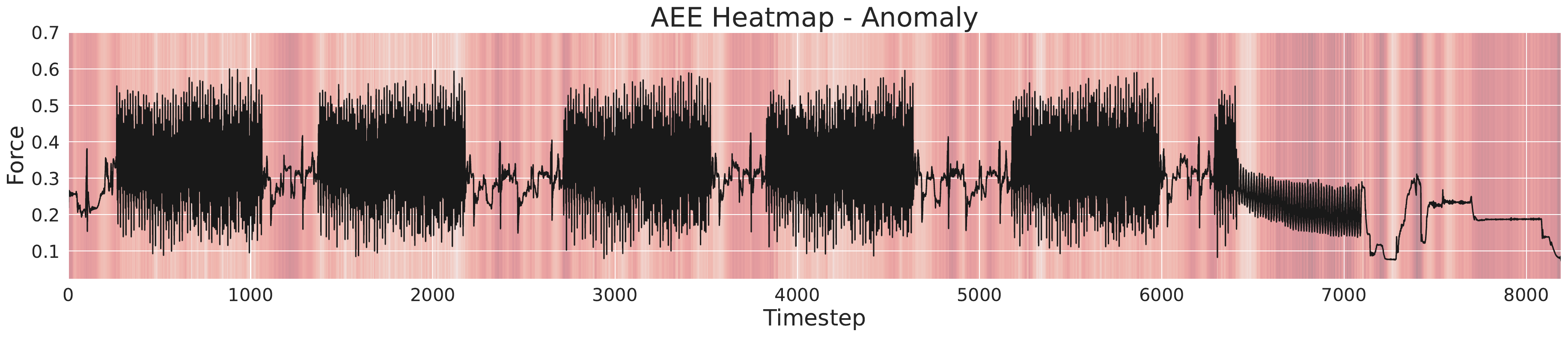}
\caption[Aggregated Heatmap]{\textbf{Aggregated Explanatory Ensemble - AEE.} 
The aggregated explanation is represented by a heat map in the background, with deeper shades of red indicating areas of greater significance for its explanation in the time series. The black curve visualizes an anomalous time series, with the explanation highlighting a disruption in the pattern between the 6300 and 7200 marks in the time series. 
} \label{Aggr_image}
\end{figure}

This work utilizes convolutional autoencoders (CAE) to compress univariate time series data for anomaly detection in an automotive manufacturing plant. A complete time series is considered an outlier if the entire sequence deviates from the expected pattern. 
The purpose of utilizing CAE is to learn specific manufacturing process features and map a time series into a low-dimensional space at its bottleneck. An unsupervised anomaly detection algorithm then uses these latent features to identify outliers. Therefore, we are interested in explaining how the encoder transformation contributes to outlier detection by employing established XAI methods (\Cref{Related}) like Grad-CAM \cite{Selvaraju_2017_ICCV}, LIME \cite{lime}, SHAP \cite{37}, and LRP \cite{1, 39, 27}, since we use the CAE's latent features for detecting outliers.
The explanations of the XAI techniques mentioned above lead to fluctuations due to their unique features.
This diversity motivates us to combine these explanations into a single, more comprehensive one: \textbf{AEE} - \textbf{A}ggregated \textbf{E}xplanatory \textbf{E}nsemble (see \Cref{MXAI}), visualized in \Cref{Aggr_image} for an anomalous time series instance. Since ground-truth data for evaluating the produced explanations are often missing, counterfactuals are widely recognized as an effective quantitative method for XAI techniques \cite{pmlr-v97-goyal19a, 10.1007/978-3-031-18840-4_37}. In our work, we implement a revised version of the quality measurement (QM) procedure, originally proposed in \cite{schlegel}, as detailed in \Cref{QM}.
We assess the effectiveness of the techniques both qualitatively and quantitatively (see \Cref{Experimentation}) based on the underlying manufacturing process. Our primary focus is on discussing their implications for erroneous time series data to gain insights into the property of being an outlier as a complete time series.

\section{Related XAI Approaches}  \label{Related}
The following section briefly mentions the XAI approaches used in this work. They were chosen for their well-known status and ability to cover explainability from different aspects, e.g., local vs. global and model-agnostic vs. model-specific explanations. These techniques share the goal of providing post hoc explanations, but each employs different approaches to achieve explainability.

\noindent
\textbf{CAM} (Class Activation Mapping), proposed by Zhou et al. \cite{Zhou}, is a local and model-specific technique for explaining convolutional neural networks (CNN). Selvaraju et al. \cite{Selvaraju_2017_ICCV} enhanced this approach with Grad-CAM, incorporating gradients into the explanation process. This improvement removes the requirement for a global average pooling layer, making the method applicable to a broader range of model architectures.

\noindent
\textbf{LIME} (Local Interpretable Model-agnostic Explanations), proposed by Ribeiro et al. \cite{lime}, is another well-known local but model-agnostic technique. It achieves interpretability by locally approximating the behavior of a complex neural network with an interpretable machine learning algorithm. 

\noindent
\textbf{SHAP} (SHapley Additive exPlanations) is a game theory-inspired model-agnostic technique proposed by Lundberg et al. \cite{37}, which can provide local or global explanations. It computes Shapley values for the model's data inputs, providing a global explanation that considers multiple data instances simultaneously. Studies carried out in \cite{SHAP_AUTO, SHAP_AE} utilized SHAP to unravel the inner workings of a complete autoencoder.

\noindent
\textbf{LRP} (Layer-wise Relevance Propagation) aims at comprehending a model's inner workings by retroactively propagating the internal values of its layers. Following this strategy, LRP seeks to assign relevance scores to different input features \cite{1, 39, 27}.

\section{Application of XAI to Autoencoder}\label{XAI}

\textbf{Notation.} 
An univariate time series instance $\mathbf{t}$ is fed into the convolutional autoencoder via the function $\hat{\mathbf{t}} = D(E(\mathbf{t}))$, with encoder $E$, decoder $D$, and latent space $L$, where $L = E(\mathbf{t})$. The output $\hat{\mathbf{t}}$ is a reconstruction of $\mathbf{t}$. We denote explanation $\mathbf{E}$ as the output of an XAI technique for a time series $\mathbf{t}$.

\subsection{Adapting XAI Techniques to Encoder} \label{encoder_xai}
We employ a 1D convolutional autoencoder (1D CAE) to reduce feature dimensions and detect anomalies in time series data (see \Cref{Experimental} for more details). We apply XAI techniques to the encoder since we use its output for anomaly detection. While the straightforward architecture facilitates the application of the XAI methods introduced in \Cref{Related}, their utilization, although widely applied in diverse machine learning scenarios, remains relatively limited within the realm of 1D CAE, mainly when applied to time series data.
We adapt these methods to improve their capability to provide 1D explanations for 1D convolutional networks in the form of heatmaps.
The application of the XAI methods above yields two distinct types of explanations:

\begin{itemize}
    \item \textbf{\textit{Individual Feature Explanation:}} For each latent feature, \( l_i \in L \) (where $i$ is the feature index), we generate a dedicated heatmap. This allows us to inspect how individual features in $\mathbf{t}$ contribute to the reconstruction process (see \Cref{Individual_heatmaps_}). 
    \item \textbf{\textit{Combined Feature Explanation:}} In addition to the individual views, we also create a unified heatmap that integrates all latent features into a single representation (see \Cref{fig:subfig1}). This combined view provides a holistic understanding of how the interplay between features in $\mathbf{t}$ influences the reconstruction process. All experiments and figures in this paper use combined feature explanations.
\end{itemize}

\subsection{AEE - \textbf{A}ggregated \textbf{E}xplanatory \textbf{E}nsemble.}\label{MXAI}
With the application of the covered XAI approaches (see \Cref{Related}), we generate a set of diverse explanations.
Each XAI technique provides distinct insights: Grad-CAM emphasizes spatial relevance, LIME offers local interpretability, SHAP delivers global explanations, and LRP traces relevance propagation (see \Cref{fig:main_1}). By aggregating these methods, AEE leverages their strengths for a holistic understanding of anomalies.
We restrict AEE to a time series $\mathbf{t}$ that stores the diverse explanations $\mathbf{E}_i$ in an array $\mathbf{E}^x_i$, where $i$ indicates the index of $\mathbf{t}$ and $x$ denotes the underlying XAI technique. 
To ensure equal consideration for each explanation, we individually scale each element $\mathbf{E}_i^x$ based on its importance scores.
Mathematically, the scaled explanation $\mathbf{SE}_i^x$ is given by:
\begin{equation}
\mathbf{SE}_i^x= \left(\frac{{\mathbf{E}_i^x - \min(\mathbf{E}^x)}}{{\max(\mathbf{E}^x) - \min(\mathbf{E}^x)}}\right) \times (a_{\text{max}} - a_{\text{min}}) + a_{\text{min}}.
\end{equation}

Here, $a_{\text{min}}$ and $a_{\text{max}}$ are the minimum and maximum values desired for scaling $\mathbf{SE}_i^x$.
After scaling, we compute the mean value for each point $i$ on the $X$ axis. We denote the aggregated version as $\mathbf{A}_i$, where $i$ represents the aggregated value for the $i$th point of the time series $\mathbf{t}$ on the $X$ axis, mathematically:

\begin{equation}
\mathbf{A}_i = \frac{1}{|x|} \sum_{j=1}^{|x|} \mathbf{SE}_i^{j}.
\end{equation}

Here, $|x|$ denotes the count of explanations stored in $\mathbf{SE}_i^x$.
Alternatively, a weighting scheme can be employed instead of equal contribution to assign more relevance to specific explanations.

\subsection{Quality Measurement of Encoder's Explanation}\label{QM}

Given the interpretability constraints of the XAI results \cite{molnar2019}, we quantitatively analyze the explanations generated by each method using a modified version of the quality measurement function proposed by Schlegel et al. \cite{schlegel}. 
In this work, the XAI techniques focus on the encoder’s explainability, resulting in a multi-regression task. Using the reconstruction error as a quality measurement would involve the decoder, misleading the measurement of the encoder’s explanation.
Instead, we aim to analyze the projections of the original time series $\mathbf{t}$, a randomly perturbed version $\mathbf{t_r^c}$, and a version perturbed based on explanation results $\mathbf{t^c}$ in the latent space.
This approach operates independently of the decoder, focusing on explaining the techniques applied to the encoder. Adversarial perturbations \cite{adversarial}, which manipulate predictions, suggest that the distance between $\mathbf{t_r^c}$ and $\mathbf{t}$ should be smaller than between $\mathbf{t}$ and $\mathbf{t^c}$.
Thus, we define the quality measurement for the encoder as:

\begin{equation} \label{equation:4}
qm_e(\mathbf{t,t}) \leq qm_e(\mathbf{t,t_r^c}) \leq qm_e(\mathbf{t,t^c}).
\end{equation}

Here, $qm_e$ measures the Euclidean distance between the original and perturbed time series in the latent space. The underlying theory is that perturbations based on explanation results have a more significant impact on the model's predictions than random noise \cite{pmlr-v97-goyal19a, 10.1007/978-3-031-18840-4_37}. The approach applies to individual and combined feature explanations, revealing the importance of features for the outlierness property of the instance.

\section{Experimentation} \label{Experimentation}

\subsection{Experimental Setup} \label{Experimental}
\textbf{Dataset.}
As introduced in \Cref{Intro}, our demonstration employs univariate time series data originating from a production plant. More specifically, it covers one process in a manufacturing line consisting of multiple processes.
The dataset consists of 18,412 time series instances, each containing 8,192 data points. The test station (end of the line) automatically labels the data to indicate normal operation (OK) or an error (not OK—NOK) during production; NOK accounts for 0.68\% overall. This information can be used to assess whether the found outliers correspond to actual errors identified by the test station.

\noindent
\textbf{Anomaly Detection Pipeline.} 
Our 1D CAE architecture comprises three convolutional layers with ReLU activation functions, followed by max-pooling layers and a bottleneck layer with a three-dimensional latent space (see \Cref{AE_appendix}).
We divide the data into three sets to train the AE: a training set for model training (0.66\% NOK), a validation set (0.74\% NOK), and a separate set for testing (0.70\% NOK) the model's performance. 

\begin{table}[t]
    \caption{\textbf{Anomaly Detection Performance Measurements.} The table contains the precision, recall, and F1-score performance metrics of the developed anomaly detection pipeline for the test set.}
    \label{AE_results}
    \centering
    \begin{tabular}{>{\centering\arraybackslash}p{1.4cm} >{\centering\arraybackslash}p{1.6cm} >{\centering\arraybackslash}p{1.5cm} >{\centering\arraybackslash}p{1.6cm} >{\centering\arraybackslash}p{1.6cm}}
    \toprule
    \textbf{Class} & \textbf{Precision} & \textbf{Recall} & \textbf{F1-Score} & \textbf{Support}\\
    \midrule
    0 (OK)& 1.00 & 1.00 & 1.00 & 5441 \\
    1 (NOK)& 0.89 & 0.63 & 0.74 & 38\\
    \bottomrule
    \end{tabular}
\end{table}

We intentionally include known anomalies in the training process, as instances with NOK labels may contain errors originating from other processes in the manufacturing line that the time series does not cover. In addition, the proportion of abnormal instances is low enough (less than 1\%) that the autoencoder continues to learn to reconstruct the time series correctly without learning anomalies. 
The pipeline consists of an anomaly detection mechanism that utilizes the latent feature space as input (see \Cref{lpp}). Specifically, we employ the density-based spatial clustering of applications with noise (DBSCAN) algorithm \cite{Reynolds2009}. \Cref{AE_results} presents the performance metrics of the anomaly detection pipeline, categorized into NOK and OK classes. These results are based on the evaluation of the test dataset.

\begin{figure}[ht] 
    \centering
    \begin{subfigure}[b]{\textwidth}
        \centering
        \includegraphics[width=\linewidth]{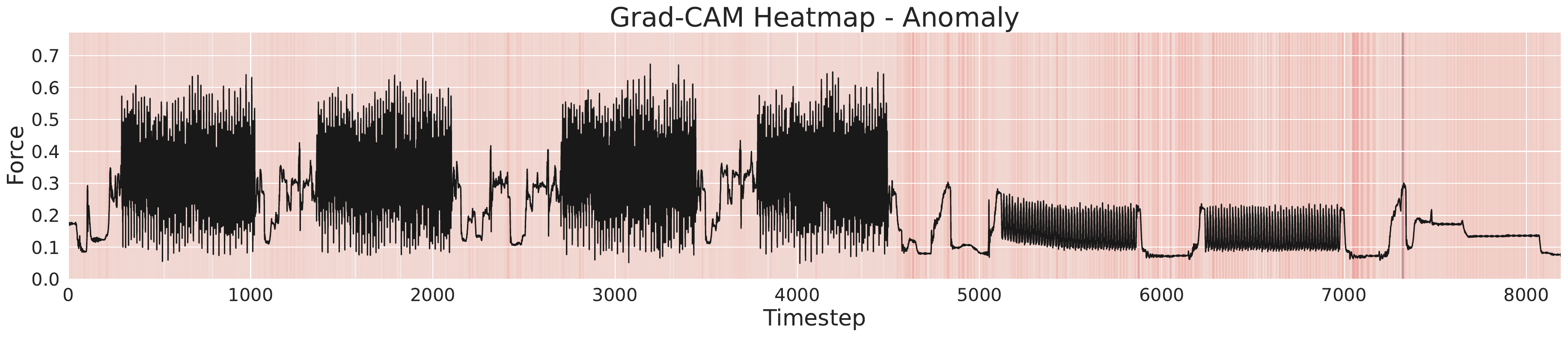}
        \caption{Grad-CAM}
        \label{fig:subfig1}
    \end{subfigure}
    \hfill
    \begin{subfigure}[b]{\textwidth}
        \centering
        \includegraphics[width=\linewidth]{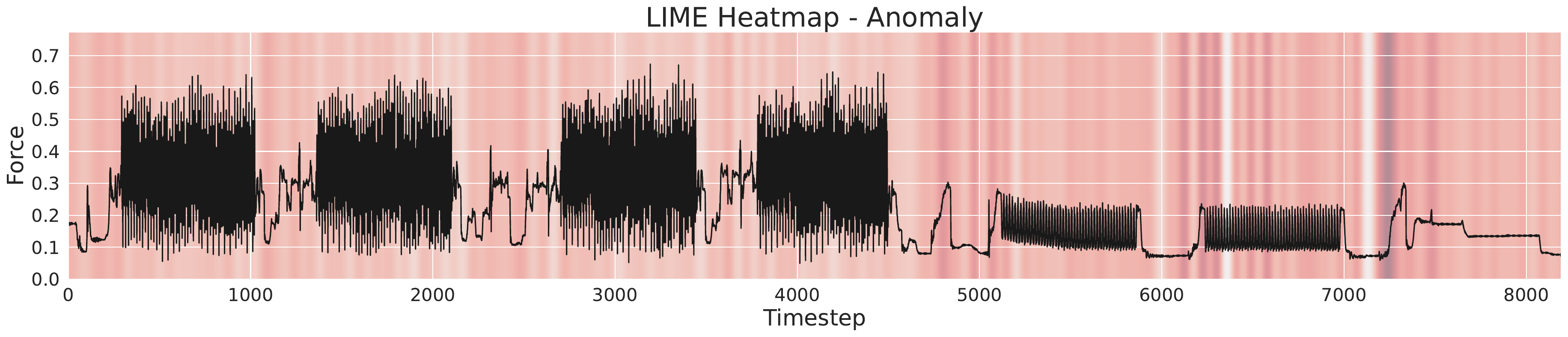}
        \caption{LIME}
        \label{fig:subfig2}
    \end{subfigure}
    \hfill
    \begin{subfigure}[b]{\textwidth}
        \centering
        \includegraphics[width=\linewidth]{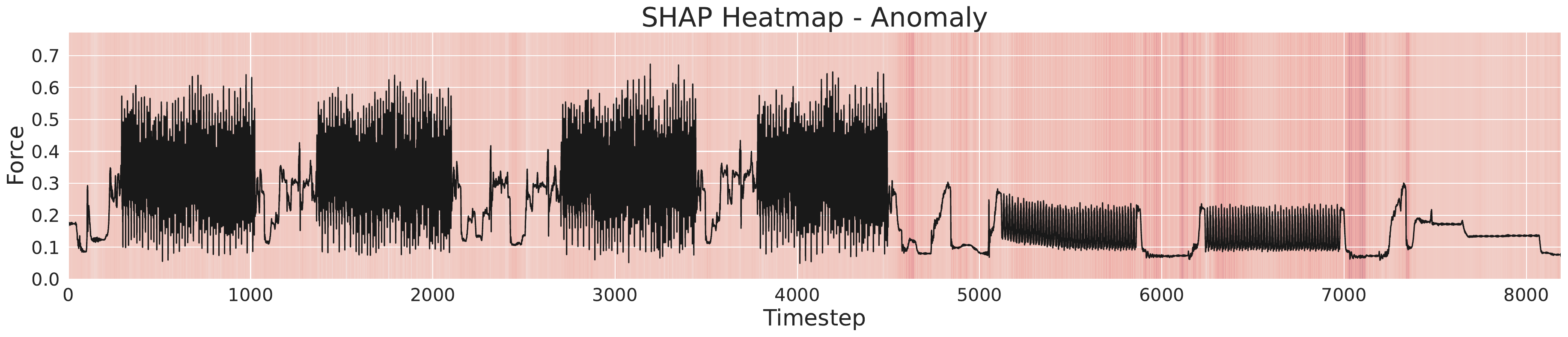}
        \caption{SHAP}
        \label{fig:subfig3}
    \end{subfigure}
    \begin{subfigure}[b]{\textwidth}
        \centering
        \includegraphics[width=\linewidth]{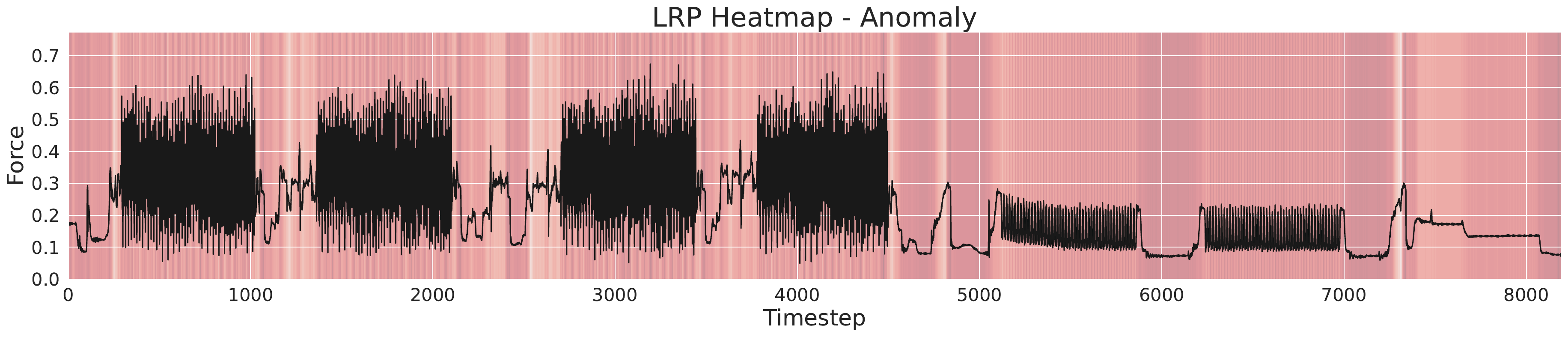}
        \caption{LRP}
        \label{fig:subfig4}
    \end{subfigure}
    \caption{\textbf{Individual XAI Results.} The XAI results are presented in the form of heatmaps. The black portions of the images denote the time series signal. These displayed instances were identified as abnormal by the AE's pipeline. The heatmap in the background indicates feature importance using varying intensities of red. We must evaluate color intensity individually as XAI techniques calculate feature importance differently.}
    \label{fig:main_1}
\end{figure}

\subsection{Qualitative Evaluation - Anomaly Interpretation} \label{quali}

\begin{figure}[h]
    \centering
    \includegraphics[width=\linewidth]{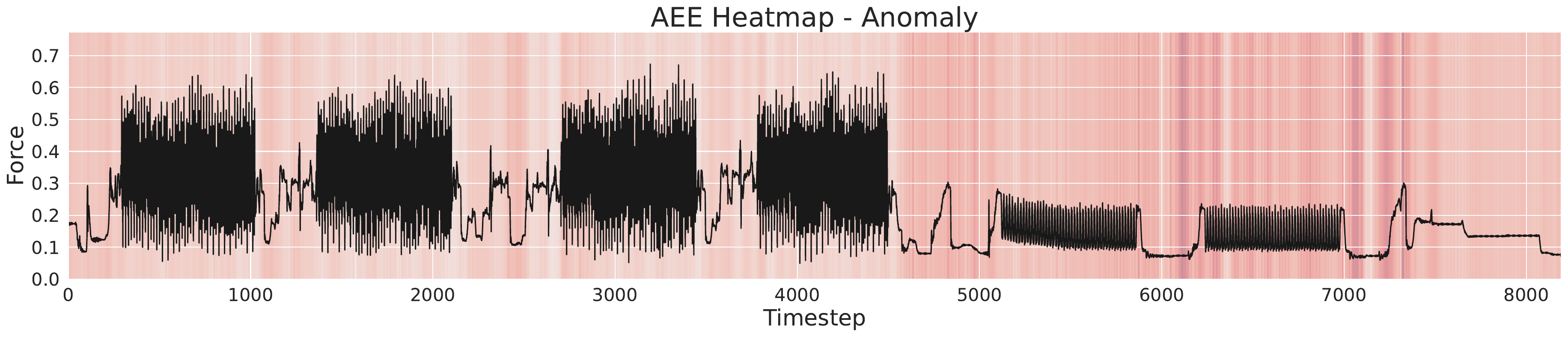}
    \caption{\textbf{AEE XAI Results.} This figure presents the results of the XAI analysis for the AEE approach. The format and layout of these explanations are consistent with those shown in \Cref{fig:main_1}.}
    \label{fig:main_2}
\end{figure}

In the following, we discuss the utility of XAI techniques to interpret the encoder, focusing on understanding why specific instances lead to anomalies by leveraging domain-specific knowledge of the underlying manufacturing process. We examine an exemplary time series classified as NOK for all explanation techniques in \Cref{fig:main_1} and \Cref{fig:main_2}. The illustrative case diverges notably in its final third segment, as the pattern is expected to exhibit distinct characteristics compared to the preceding two thirds of the time series (see \Cref{ok_series}). We explicitly demonstrate the anomalies shown here using examples that are easy to visually understand as outsiders. 

We initiate with Grad-CAM (\Cref{fig:subfig1}), revealing a heatmap that distinctly accentuates positions later in $\mathbf{t}$, precisely aligning with observable areas of technical failures in the manufacturing process. This targeted explanation effectively identifies the specific region preceding real-world anomalies.
Subsequently, LIME (\Cref{fig:subfig2}) highlights the same area as CAM, but its interpretation is more straightforward because of its apparent intensity. Moreover, it also subtly indicates regions in intermediate areas of the time series.
SHAP (\Cref{fig:subfig3}) pinpoints the same critical area of primary importance, consistent with the findings of the previous methods.
Compared to the preceding, the final standalone method LRP (\Cref{fig:subfig4}) diverges in its explanation. Although it does not explicitly emphasize the most pronounced pattern, it assigns varying degrees of importance to different segments and provides valuable insights for manual analysis by a domain expert.

\Cref{fig:main_2} shows the aggregated explanation. Parallel to Grad-CAM and SHAP, the region signaling an abnormal pattern is precisely accentuated, and the aggregated version amplifies the color representation, enhancing interpretability.
Besides confirming the importance of the known area, this approach offers additional insights into other parts of the time series, e.g., it prioritizes early regions that indicate possible technical abnormalities. Repeated experiments prove its explanations are more stable due to its aggregation property, mitigating the negative implications of instability \cite{DBLP:journals/corr/abs-2006-05714}. 

\subsection{Quantitative Evaluation - XAI Quality Measurement.} \label{quanti}
\begin{figure}[h]
\centering
\includegraphics[width=\linewidth]{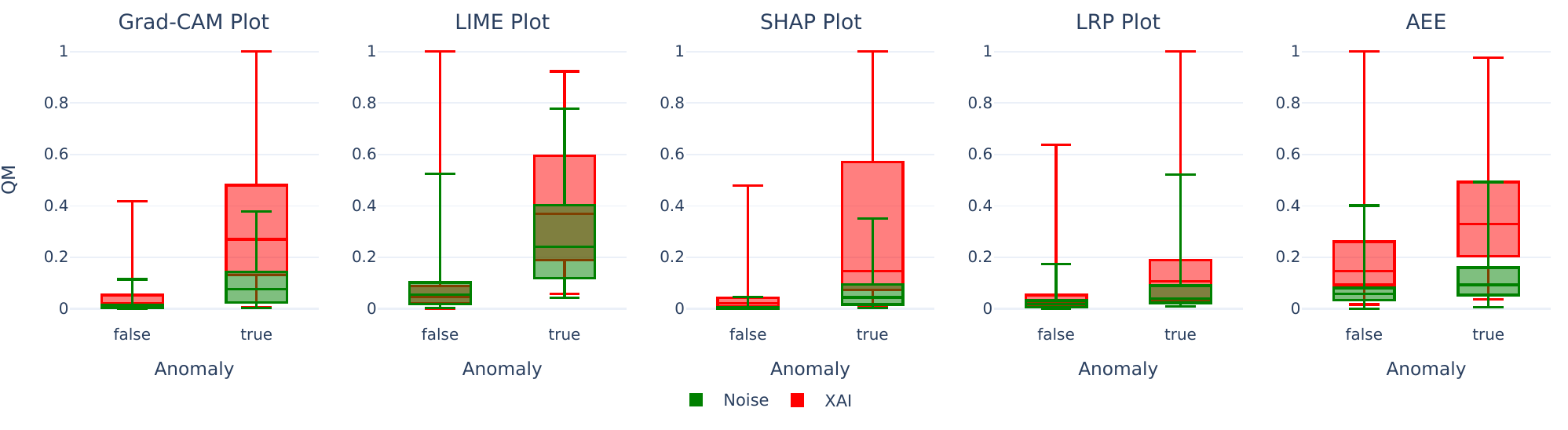}
\caption[Interquartile range quality measurements.]{\textbf{Interquartile Range Quality Measurements.} The visualization depicts quality measurement scores for each XAI technique, categorized into true anomalies (NOK) and false anomalies (OK). The measurements are further stratified into noise ($\mathbf{t^c_r}$ - XAI shuffled), denoted by green, and XAI ($\mathbf{t^c}$ - XAI perturbated), represented by red.}
\label{eval_image}
\end{figure}

\Cref{eval_image} depicts the QM (normalized Euclidean distances) distributions, where boxes represent the interquartile range (IQR) from Q1 to Q3, with a median line (Q2). The fences extend \textpm 1.5 times the IQR. The OK category includes 100 randomly selected instances, and the NOK category comprises 38. The noise/shuffle box (green) represents QM values $\mathbf{t^c_r}$, and the XAI box (red) represents $\mathbf{t^c}$.

The visualization indicates that each QM XAI score consistently outperforms its QM noise counterpart. The scores for the NOK cluster are significantly higher, demonstrating the effectiveness of using explanations for outlier interpretation. LRP and LIME overlap between Noise and XAI, while Grad-CAM and SHAP display a clearer separation in their explanations. The AEE produces a significant result, indicating that aggregating multiple explanations sharpens the distinction between relevant and irrelevant features within a time series, improving explanation quality.

\subsection{Limitations and Future Work} \label{limits}

Our study applied XAI techniques to CAE, leaving the potential for other architectures such as variational autoencoders (VAE) \cite{park2018multimodal} and recurrent neural networks (RNN) \cite{3, 8, s23052844} unexplored. Additionally, the evaluation of these techniques was primarily based on qualitative assessments, as anomalies required examination by domain experts. Future research on datasets not requiring expert knowledge should consider integrating additional quantitative methods to complement qualitative insights \cite{anecdotal}.
In addition, a clear distinction between explanation and interpretation should be established \cite{BARREDOARRIETA202082}, recognizing that not all explanations are inherently human-interpretable \cite{10.1145/3387166}, as it was sometimes the case in this scenario.

Furthermore, exploring different weighting schemes for AEE could enhance the interpretation and accuracy of feature importance calculations in various scenarios. 
We outlined the experimentation on the time series manufacturing use case. Future research could involve testing the AEE approach across various data types, such as images or text. Regarding XAI approaches, future work could focus on improving time series segmentation using foundation models \cite{knab2024dseglime}, particularly beneficial for LIME. Another promising direction is to direct the explanations not toward the latent features themselves but the classes in the latent space that signify the presence or absence of anomalies.
Lastly, extending this methodology to multivariate time series \cite{8411269, 8, multivariate, Segal} or even multimodal data \cite{park2018multimodal} presents another intriguing avenue for future exploration.

\section{Conclusion} \label{Conclusions}
This paper contributes to the application of XAI techniques to CAEs for analyzing outlier properties within the latent space of time series data in the operational context of a manufacturing plant. We employed well-established XAI methods to demonstrate the practicality and effectiveness of these techniques in interpreting outliers. In addition, we introduced AEE, an ensemble of multiple XAI techniques. We quantitatively evaluated the different explanations using a QM approach specifically modified to fit the encoder of an AE.
Moreover, the application of XAI techniques provided explanations for these outliers, accurately highlighting the abnormal segments within the time series. This alignment confirms the utility of XAI in providing meaningful insights into anomalies and building confidence in the system through the interpretation of XAI results.

\section*{Acknowledgments}
This work was supported by the German Federal Ministry for Economic Affairs and Climate Action (BMWK).

%
%
%
%
\bibliography{thesis-ref}

\newpage
\appendix

\section{Individual Feature Explanation} \label{Individual_heatmaps_}
\begin{figure}[ht] 
    \centering
    \begin{subfigure}[b]{\textwidth}
    \caption{Heatmap Latent Feature One}
            \includegraphics[width=\linewidth]{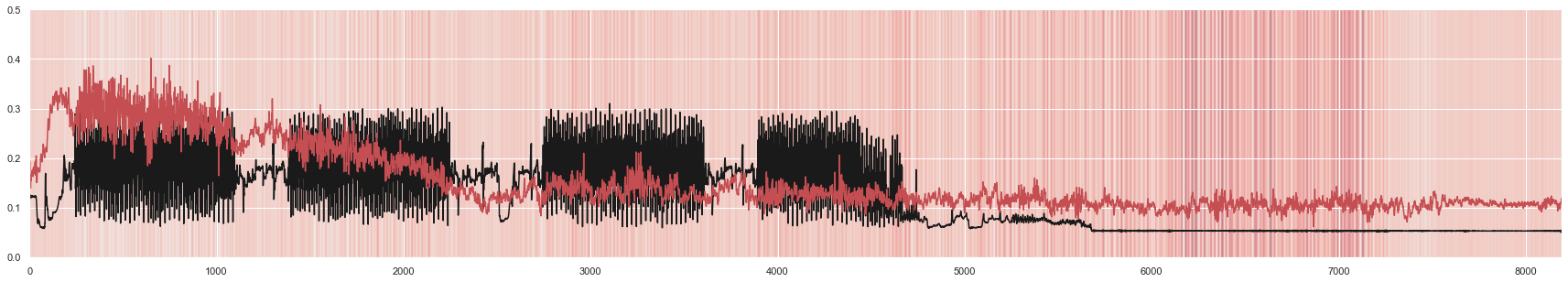}
    \end{subfigure}
    \begin{subfigure}[b]{\textwidth}
    \caption{Heatmap Latent Feature Two}
            \includegraphics[width=\linewidth]{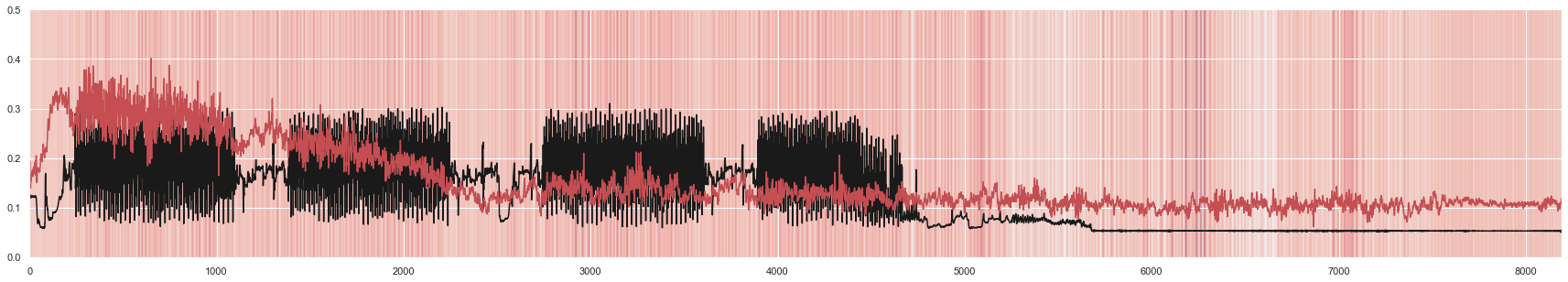}
    \end{subfigure}
    \begin{subfigure}[b]{\textwidth}
    \caption{Heatmap Latent Feature Three}
            \includegraphics[width=\linewidth]{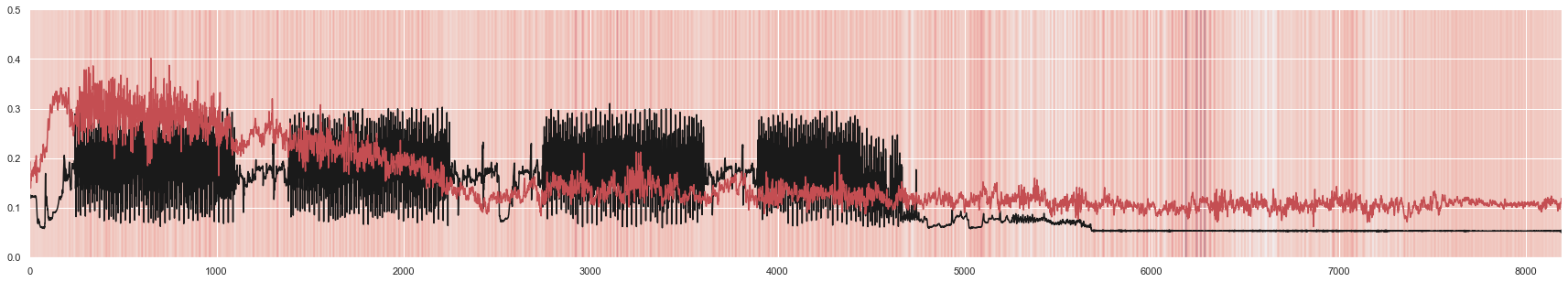}
    \end{subfigure}
\caption[gr]{\textbf{Grad-CAM - Individual Feature Heatmaps.} The images illustrate individual latent feature explanations in the form of a heatmap generated by Grad-CAM. The black curve illustrates the original time series, while the red curve represents its reconstruction.}
\label{Individual_heatmaps}
\end{figure}

\Cref{Individual_heatmaps} shows an instance that the pipeline classified as NOK, featuring the reconstructed time series in red and the original time series in black. The underlying explanation is provided through individual feature explanations, where a distinct heatmap visually explains each latent feature.

\section{Autoencoder Architecture} \label{AE_appendix}

\begin{figure}[ht]
\centering
\includegraphics[width=\linewidth]{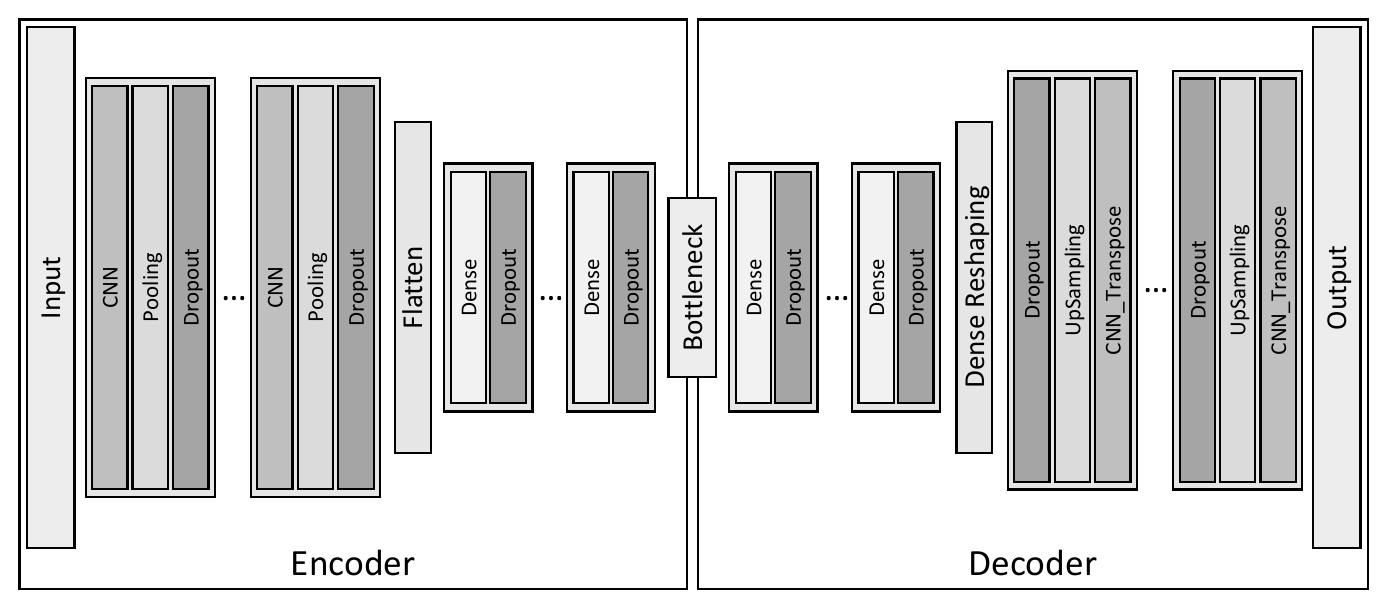}
\caption[The Building Blocks of an Autoencoder: An Abstract Architecture]{\textbf{The Building Blocks of an Autoencoder: An Abstract Architecture.} The AE's architecture comprises diverse blocks, each possessing unique internal attributes and dimensions. As a result, the encoder and decoder are constructed separately, deviating from the conventional symmetrical autoencoder. These blocks encompass convolutional layers and their associated operations alongside a dense block that integrates dense and dropout layers.}
\label{ref_ae}
\end{figure}
\noindent
In the following, we present the defined search space of hyperparameters for tuning an AE in this work. The search space has been explored by 100 runs and 500 epochs each. \Cref{ref_ae} represents the building blocks we tuned during this process.

\begin{itemize}
\item The amount of \textbf{CNN blocks} consists of an optional dropout and max pooling layer. We restrict this size to at least one and a maximum of three blocks. This number applies to both the encoder and the decoder. 
\item In contrast to the number of CNN blocks, this number varies for \textbf{DNN blocks} between the encoder and decoder parts. Both can have up to two DNN blocks.  
\item Each convolutional layer is optimized with a specific number of \textbf{filters} in its operation, namely 16, 32, 64, and 128. Furthermore, the \textbf{kernel size} is tuned to either 8, 16, or 32. While it is possible to consider additional values for these parameters, doing so would increase the search space for the tuner.
\item The \textbf{dropout layer} is optional for each CNN and DNN block. Possible \textbf{dropout rates} range is 0.1, 0.2, 0.3, 0.4, and 0.5. 
\item \textbf{Max pooling} is another optional layer in the CNN-Block but with a fixed pooling size of two. 
\item The range of \textbf{neurons} in a dense layer is 32, 64, 128, and 256, respectively. 
\item The \textbf{activation function} chosen for each layer in the autoencoder remains consistent, namely, the ReLu, Tanh, Sigmoid, or Softmax function. However, only the output layer of the decoder is individually tailored to these four functions.
\end{itemize}

\section{Latent Space Plot} \label{lpp}

\begin{figure}[h]
\centering
\includegraphics[width=\linewidth]{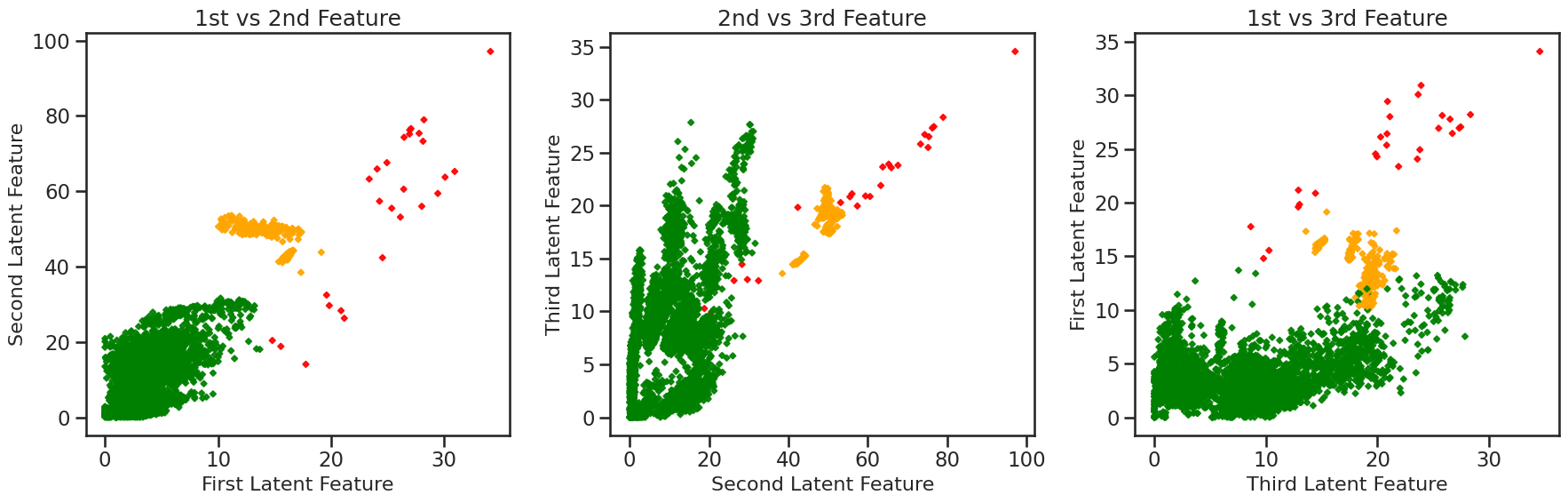}
\caption[Final AE Latent Space Plot]{\textbf{Latent Space Visualization.} Two-dimensional latent space representation of the autoencoder's features. Green and orange points represent instances assigned to two distinct clusters, while red points are identified as outliers. }
\label{lpp_figure}
\end{figure}

The encoder's output projection is shown in \Cref{lpp_figure}. This figure displays the latent variables on a two-dimensional scale for easier interpretation. Each point on the plot corresponds to a mapped instance, representing a complete time series from the test dataset. The colors indicate the categorization by DBSCAN in the latent space: red points are outliers, orange points represent instances with manually detected deviations yet considered OK, and green points indicate cases with no apparent deviations, also classified as OK.

\section{Exemplary Non-Outlier Time Series} \label{ok_series}

\begin{figure}[ht]
\centering
\includegraphics[width=\linewidth]{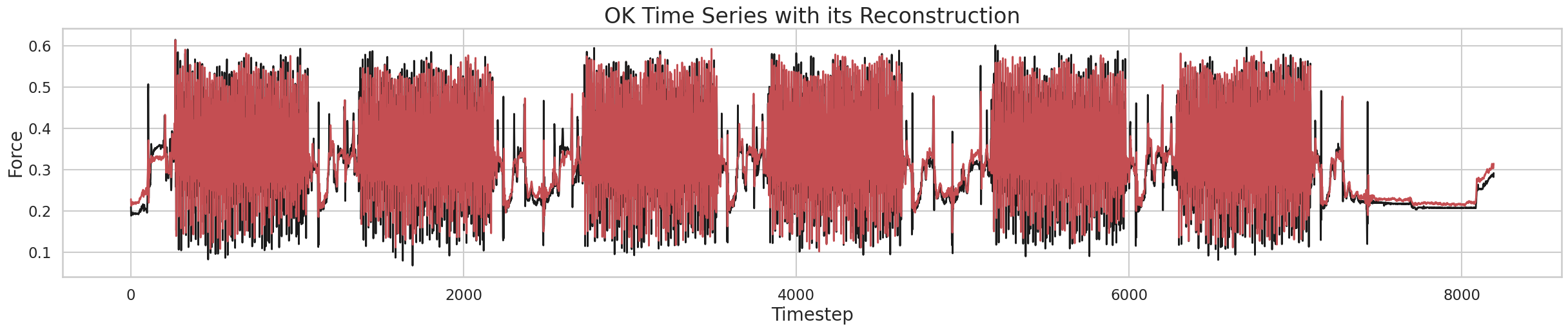}
\caption[Final AE Latent Space Plot]{\textbf{Time Series Reconstruction.} The figure illustrates a time series, depicted in black, classified as OK. The corresponding reconstruction through the autoencoder is shown in red. }
\label{ok_time_series}
\end{figure}

\Cref{ok_time_series} displays an exemplary time series classified as a non-outlier alongside its reconstruction by the autoencoder. The image demonstrates that the AE can meaningfully reconstruct the input time series. Additionally, the pattern of this time series is typical for an instance without apparent errors in this dataset.

\end{document}